\documentclass[letterpaper, 10 pt, conference]{ieeeconf}  

\IEEEoverridecommandlockouts                              

\usepackage{blindtext}
\usepackage{graphicx}

\usepackage[font=small]{caption}
\usepackage{amssymb}
\usepackage[cmex10]{amsmath}
\usepackage{float}
\usepackage{subcaption}
\usepackage{ tipa }
\usepackage{hyperref}
\usepackage[table]{xcolor}
\usepackage{mathpazo} 
\usepackage[final]{pdfpages}
\usepackage{csquotes} 
\usepackage{cite} 
\usepackage{siunitx} 
\usepackage{scrextend} 
\usepackage{courier} 
\usepackage{placeins} 
\usepackage{pgffor} 
\usepackage{csquotes}
\usepackage{tabularx}
\usepackage[english]{babel} 
\usepackage{blindtext}
\usepackage{array} 
\usepackage{mdwmath} 
\usepackage{eqparbox} 
\renewcommand{\thetable}{\arabic{table}}

\let\labelindent\relax 
\usepackage{enumitem}

\usepackage{multicol} 
\usepackage{etoolbox} 
\usepackage{hyperref} 
\usepackage{flushend}
\makeatletter
\let\NAT@parse\undefined
\makeatother

\newcommand{\vect}[1]{\boldsymbol{#1}} 

\begin{document}
	%
	
	\title{Rebellion and Obedience: The Effects of Intention Prediction in Cooperative Handheld Robots}
	%
	%
	%
	\author{Janis Stolzenwald and Walterio W. Mayol-Cuevas \thanks{ 
			Department of Computer Science, University of Bristol, UK,
			{\color{white}....} 
			janis.stolzenwald.2015@my.bristol.ac.uk, wmayol@cs.bris.ac.uk
		}
	}
	\maketitle
	%
	
	\begin{abstract}
		
		Within this work, we explore intention inference for user actions in the context of a handheld robot setup. Handheld robots share the shape and properties of handheld tools while being able to process task information and aid manipulation. Here, we propose an intention prediction model to enhance cooperative task solving. The model derives intention from the user's gaze pattern which is captured using a robot-mounted remote eye tracker. The proposed model yields real-time capabilities and reliable accuracy up to \SI{1.5}{\s} prior to predicted actions being executed. We assess the model in an assisted pick and place task and show how the robot's intention obedience or rebellion affects the cooperation with the robot.
		
		
		
	\end{abstract}
	
	

	%
	\IEEEpeerreviewmaketitle

	\section{Introduction}
	A Handheld robot shares properties of a handheld tool while being enhanced with autonomous motion as well as the ability to process task-relevant information and user signals. 
	Earlier work in this field explored the communication between user and robot to improve cooperation \cite{GreggSmith:2015bh} \cite{GreggSmith:2016hn}. Such one-way communication of task planning, however, is limited in that the robot has to lead the user. But as users exert their will and decisions, task conflicts emerge which in turn inflict user frustration and decrease cooperative task performance.
	
	As a starting point of addressing this problem, extended user perception can be introduced to allow the robot to estimate the user's point of attention via eye gaze in 3D space during task execution\cite{Stolzenwald:2018un}. An estimate of users' visual attention informs the robot about areas of users' interest. While introducing attention was preferred, particularly for temporal demanding tasks, it is still limiting. What is necessary is a model that goes beyond where the user is attending to but rather what is the user intending to do. A model of intention would allow the robot to infer the user's goal in the proximate future and go beyond reacting to immediate decisions only.
	
	Intention inference has caught researcher's attention in recent years and promising solutions have been achieved through observing user's eye gaze \cite{Huang:2016dj}, body motion \cite{Ravichandar:2015ii} or task objects \cite{Liu:2015km}. These contributions target safe interactions between humans and sedentary robots with shared workspaces. Thus, the question remains open whether there is a model which suits the setup of a handheld robot which is characterised by close shared physical dependency and a \textit{working together} rather than a \textit{turn taking} cooperative strategy.
	
\begin{figure*}[h]
		\vspace{0.5em}
		\centering
		\includegraphics[width=0.95\linewidth]{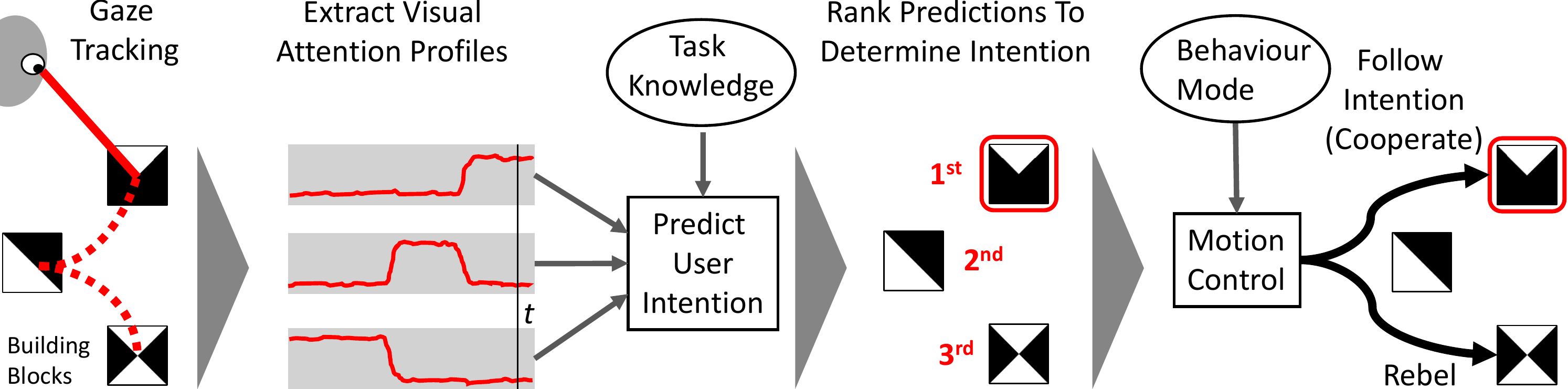}
		\caption{Overview of the intention prediction model and its use for the robot's motion control.}
		\label{fig:intentionpredictionflowchart}
	\end{figure*}

	Our work is guided by the following research questions
	\begin{enumerate}[label=\textbf{Q\arabic*}]
		\item How can user intention be modelled in the context of a handheld robot task?
		\label{Q1}
		\item To what extent does intention prediction affect the cooperation with a handheld robot?
		\label{Q2}
	\end{enumerate}
	\begin{figure}[t]
		\centering
		\includegraphics[width=0.99\linewidth]{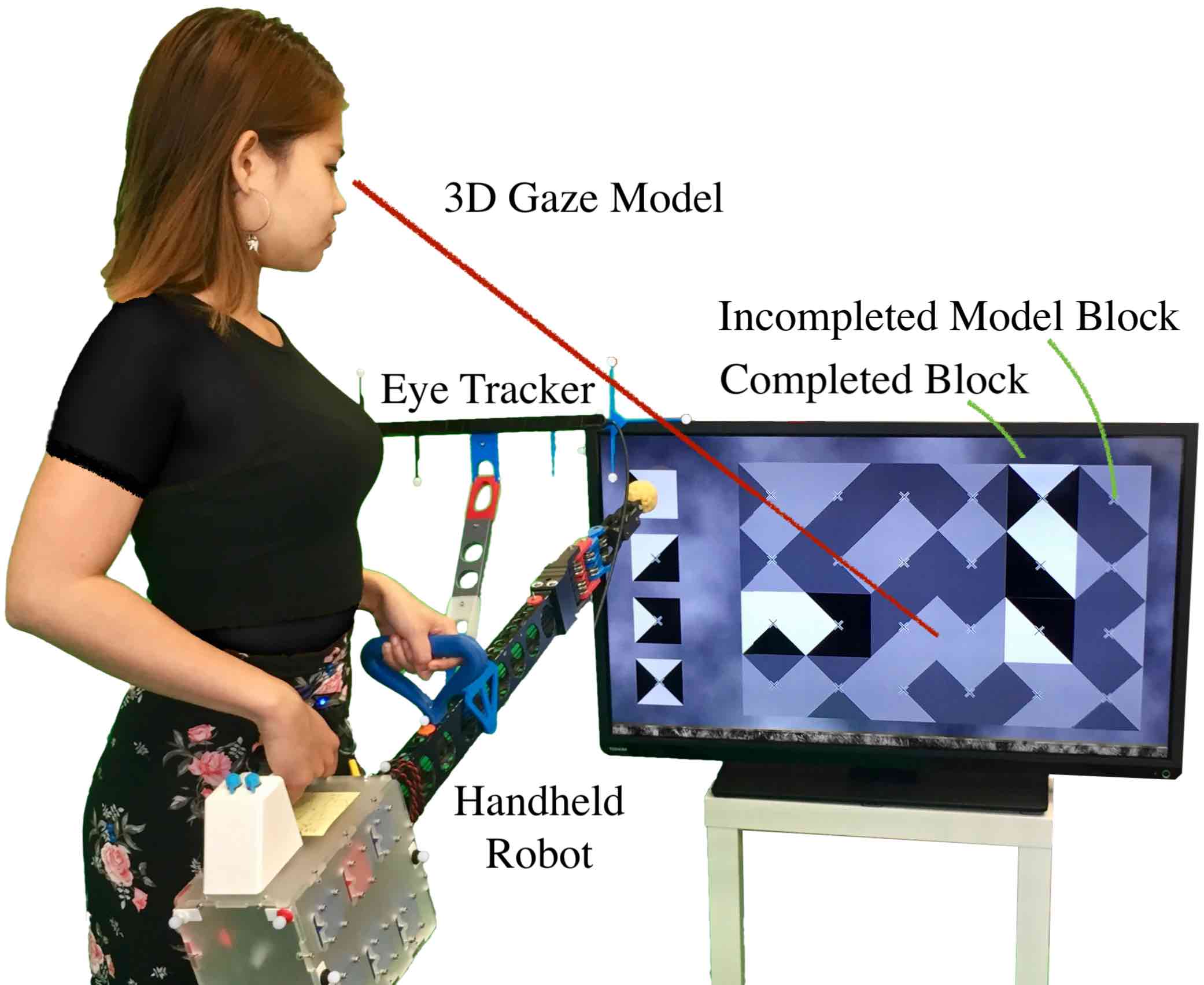}
		\caption{This picture shows a participant within our user intention prediction study. The participant uses the robot to solve an assembly task and is about to decide where to place the currently held block. Using the eye tracker the prediction system extracts the user's gaze pattern which is used for action prediction.
		}
		\label{fig:blockcopygameparticipant}
		\vspace{-0.5em}
	\end{figure}
	For our study, we use the open robotic platform\footnote{3D CAD models available from handheldrobotics.org}, introduced in \cite{GreggSmith:2016cz} in combination with an eye tracking system as reported in \cite{Stolzenwald:2018un}. Within a simulated assembly task, eye gaze information is used to predict subsequent user actions. The two principal parts of this study consist of modelling user intention in the first place followed by testing it through an assistive pick and place task. Our contribution is an intention prediction model with real-time capabilities that allows for human-robot collaboration through online plan adaptation in assistive tasks.    Figure \ref{fig:intentionpredictionflowchart} shows an overview of our proposed system.
	
	\section{Background and Related Work}
	In this section, we deliver a summary of earlier work on handheld robots and its control based on user perception. Furthermore, we review existing methods for intention inference with a focus on human gaze behaviour.
	
	\subsection{Handheld Robots}
	Early handheld robot work \cite{GreggSmith:2015bh} used a trunk-shaped robot with 4-DoF to explore issues of autonomy and task performance. This was later upgraded to a 6-DoF (joint space) mechanism \cite{GreggSmith:2016cz} and used gestures, such as pointing, to study user guidance. These earlier works demonstrate how users benefit from the robot's quick and accurate movement while the robot profits from the human's tactical motion. Most importantly, increased cooperative performance was measured with an increased level of the robot's autonomy. It was furthermore found that cooperative performance significantly increases when the robot communicates its plans e.g. via a robot-mounted display \cite{GreggSmith:2016hn}. 
	
	Within this series of work, another problem was identified: the robot does not sense the user's intention and thus potential conflicts with the robot's plan remain unsolved. For example, when the user would point the robot towards a valid subsequent goal, the robot might have already chosen a different one and keep pointing towards it rather than adapting its task plan. This led to irritation and frustration in users on whom the robot's plan was imposed on. 
	
	Efforts towards involving user perception in the robot's task planning were made in our recent work on estimating user attention\cite{Stolzenwald:2018un}. The method was inspired by work from Land et al. on how human's eye gaze is closely related to manual actions \cite{Land:2016kw}. The attention model measures the current visual attention to bias the robot's decisions. In a simulated \textit{space invader} styled task, different levels of autonomy were tested over varying configurations of speed demands. It was found that both the fully autonomous mode (robot makes every decision) and the attention driven mode (robot decides based on gaze information) outperform manual task execution. Notably, for high-speed levels, the increased performance was most evident for the attention-driven mode which was also rated more helpful and perceived rather cooperative than the fully autonomous mode. 
	
	As opposed to an intention model, the attention model would react to the current state of eye gaze information only, rather than using its history to make predictions about the user's future goals. We suggest that this would be required for cooperative task solving for complex tasks like assembly where there is an increased depth of subtasks. 

	\subsection{Intention Prediction}
	
	Intention estimation in robotics is in part driven by the demand for safe human-robot interaction and efficient cooperation. 
	
	Ravichandar et al. investigated intention inference based on human body motion. Using Microsoft Kinect motion tracking as an input for a neural network, reaching targets where successfully predicted within an anticipation time of approximately \SI{0.5}{s} prior to the hand touching the object\cite{Ravichandar:2015ii}. Similarly, Saxena et al. introduced a measure of affordance to make predictions about human actions and reached  84.1\%/74.4\% accuracy \SI{1}{s}/\SI{3}{s} in advance, respectively\cite{Koppula:2016ja}. Later, Ravichandar et al. added human eye gaze tracking to their system and used the additional data for pre-filtering to merge it with the existing motion-based model \cite{Ravichandar:2016th}. The anticipation time was increase to \SI{0.78}{s}.
	
	Huang et al. used gaze information from a head-mounted eye tracker to predict customers' choices of ingredients for sandwich making. Using a support vector machine (SVM), an accuracy of approximately 76\% was achieved with an average prediction time of \SI{1.8}{s} prior to the verbal request \cite{Huang:2015iw}.    
	In subsequent work, Huang \& Mutlu used the model as a basis for a robot's anticipatory behaviour which led to more efficient collaboration compared to following verbal commands only \cite{Huang:2016dj}.
	
	
	We note that the above work targets intention inference purposed for \textit{external} robots which are characterised by a shared workspace with a human but can move independently. It is unclear whether these methods are suitable for close cooperation as it can be found in the handheld robot setup.

	\subsection{Human Gazing Behaviour}
	The intention model presented in this paper is mainly driven by eye gaze data. Therefore, we review work on human gaze behaviour to inform the underlying assumptions of our model. 
	
	Land et al. found that fixations towards an object often precede a subsequent manual interaction by around \SI{0.6}{s} \cite{Land:2016kw}. Subsequent work revealed that the latency between eye and hand varies between different tasks \cite{Land:2001hl}. Similarly, Johansson et al. \cite{Johansson:2001ck} found that objects are most salient for human's when they are relevant for tasks planning and preceding saccades were linked to short-term memory processes in \cite{Mennie:2006fo}. 
	
	The purpose of preceding fixations in manual tasks was furthermore explored through virtual \cite{Ballard:1995iy} and real \cite{Pelz:2001fb} block design tasks. The results show that humans gather information through vision \textit{just in time} rather than memorising e.g. all object locations. 
	
	\begin{figure}[b]
		\vspace{-0.99em}
		\centering
		\includegraphics[width=0.99\linewidth]{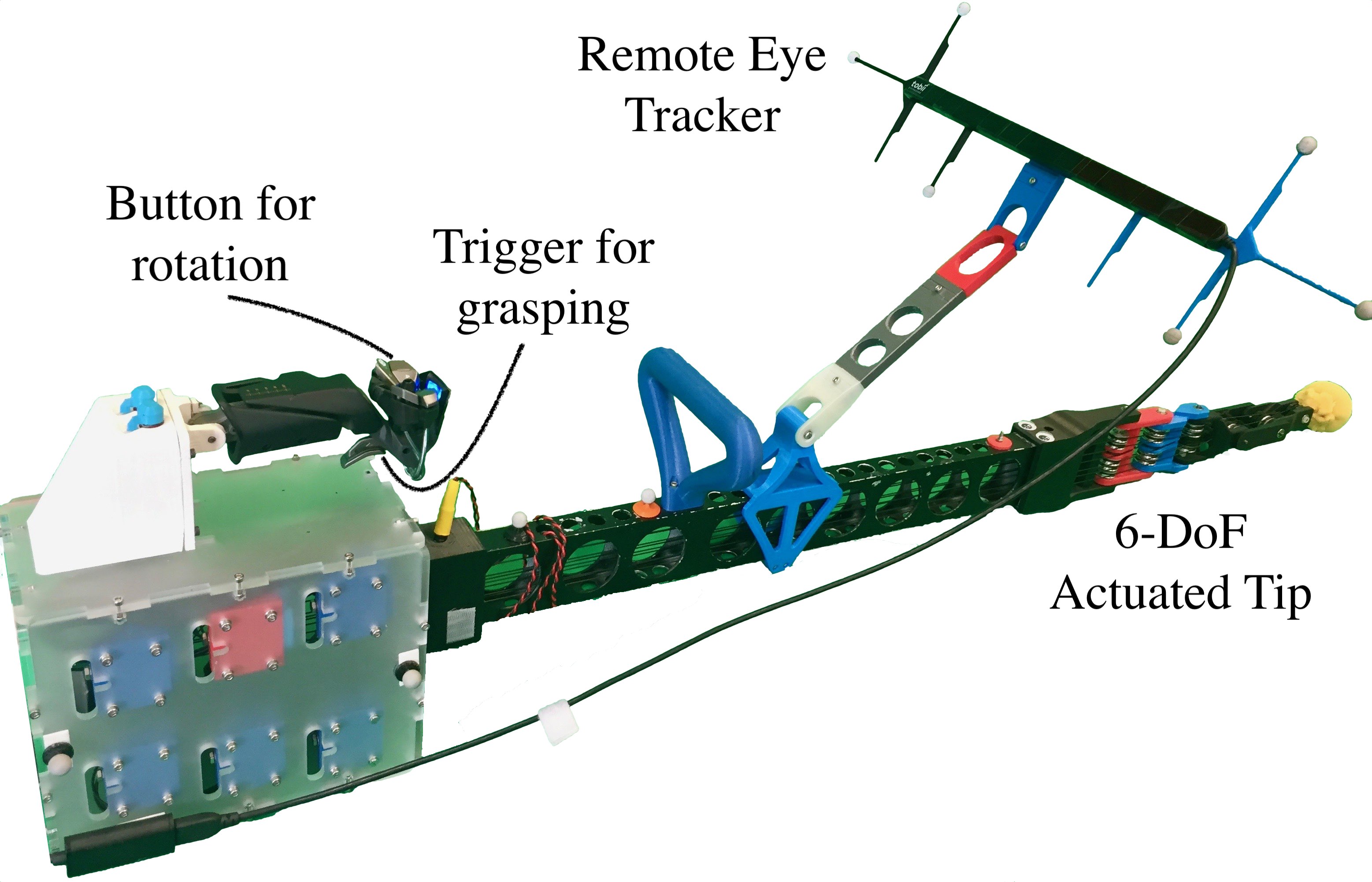}
		\caption{The handheld robot used for our study. It features a set of input buttons and a trigger at the handle, a 6-DoF tip and user perception through gaze tracking as reported in \cite{Stolzenwald:2018un}.}
		\label{fig:frontprofilerobotlabeled}
		\vspace{0.8em}
	\end{figure}
	
	\section{Prediction of User Intention}
	In this section, we describe how intention prediction is modelled for the context of a handheld robot on the basis of an assembly task. 

	\subsection{Data Collection}
\begin{figure}[h]
		\vspace{0.5em}
		\centering
		\includegraphics[width=0.99\linewidth]{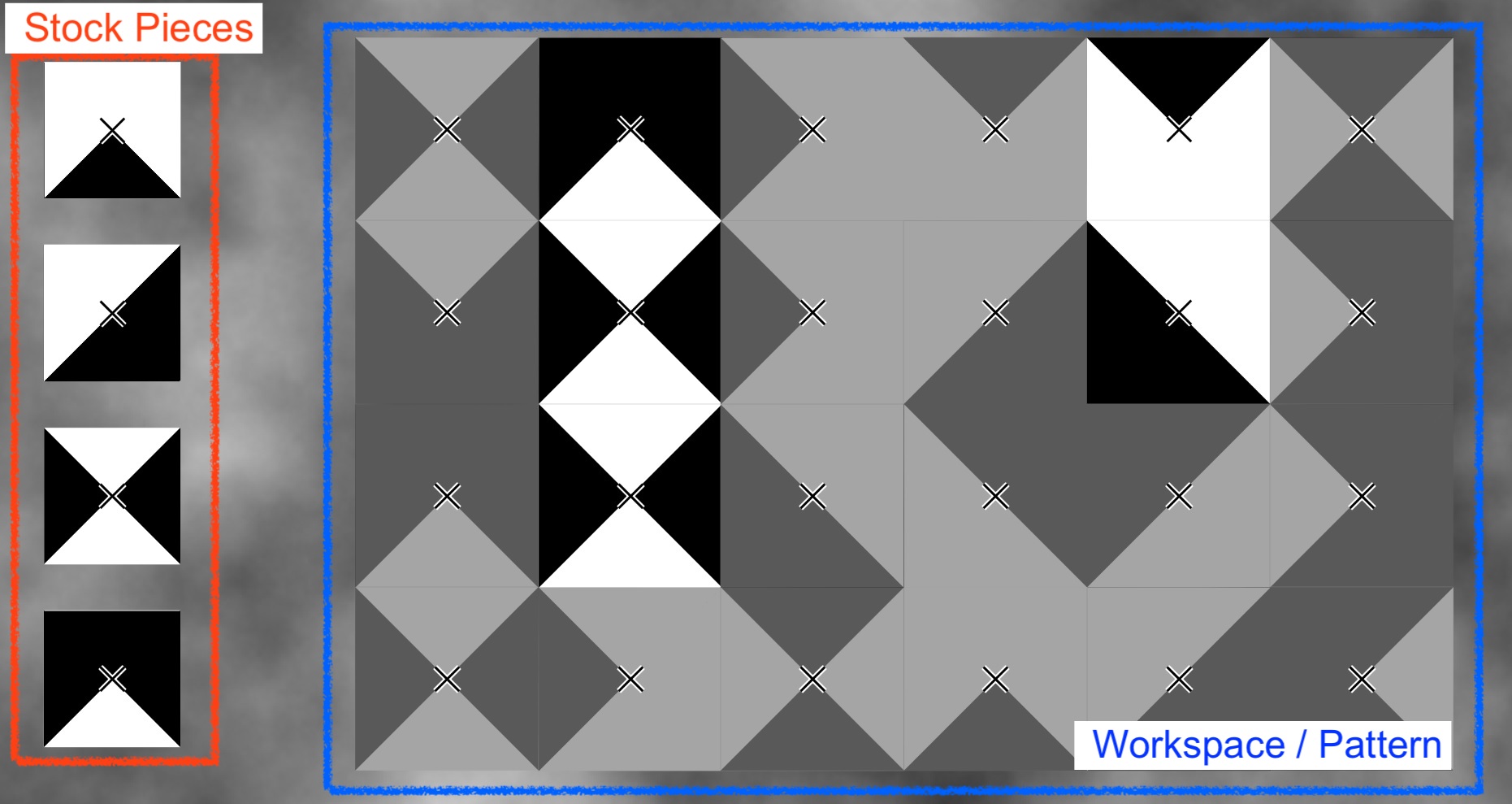}
		\caption{Layout of the block copy task on a TV display. The area is divided into stock (red) and workspace (blue). The shaded pattern pieces in the workspace area have to be completed by placing the associated pieces from the stock using the real robot.}
		\label{fig:blockcopyinitexampleareas}
	\end{figure}

\begin{figure}[h]
		\centering
		\includegraphics[width=0.99\linewidth]{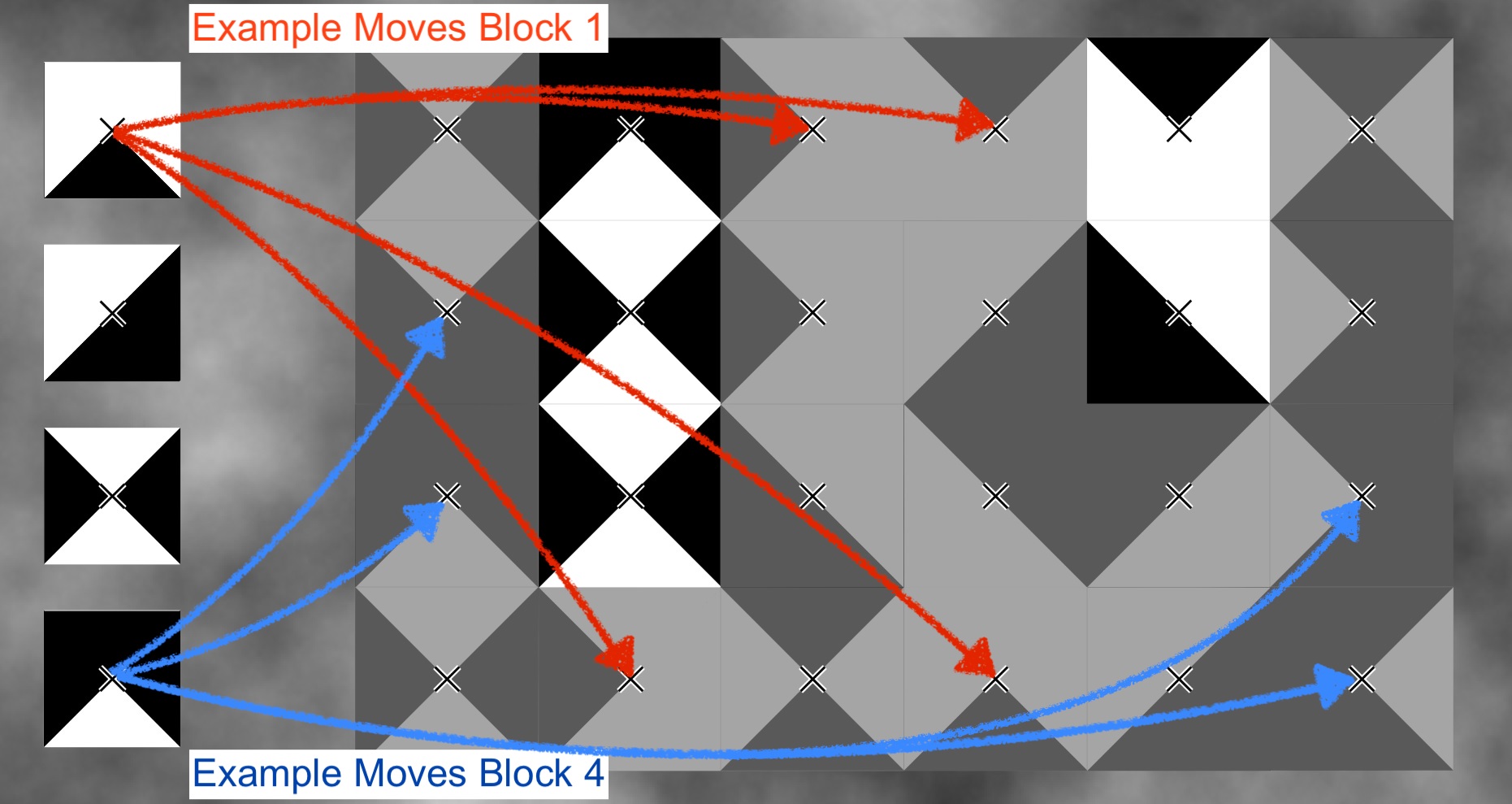}
		\caption{Examples of possible moves for block 1 and 4. A stock piece has to be moved to an associated piece in the pattern and match the model's orientation to complete it.}
		\label{fig:blockcopyinitexamplemoves}
		\vspace{-0.5em}
	\end{figure}

	We chose a simulated version of a block copying task which has been used in the context of work in hand-eye coordination \cite{Ballard:1995iy,Pelz:2001fb}. Participants of the data collection trials were asked to use the handheld robot (cf. figure \ref{fig:frontprofilerobotlabeled}) to pick blocks from a stock area and place them in the workspace area at one of the associated spaces indicated by a shaded model pattern. The task was simulated on a \SI{40}{inch} LCD TV display and the robot remained motionless during the data collection task to avoid distraction. We drew inspiration from a block design IQ test \cite{Miller:2009ib} and decided to use black and white patterns instead of colours. That way, a match with the model would, in addition, depend on the block's orientation which adds further complexity.
	An overview of the task can be seen in figure \ref{fig:blockcopyinitexampleareas}, figure \ref{fig:blockcopyinitexamplemoves} shows examples of possible picking and placing moves. 
	
	
	In order to pick or place pieces, users have to point the robot's tip towards and close to the desired location and pull/release a trigger in the handle. The position of the robot and its tip is measured via a motion tracking system\footnote{Opti Track: https://optitrack.com}\hspace{-0.4em}. The handle houses another button which can be used to rotate the grabbed piece.
	The opening or closing process of the virtual gripper takes \SI{1.3}{s} which is animated in the screen. If the participant tries to place a mismatch, the piece goes back to the stock and has to be picked up again. Participants are asked to solve the task swiftly and it is completed when all model pieces are copied. Throughout the task execution, we kept track of the user's eye gaze using a robot-mounted remote eye tracker in combination with a 3D gaze model from \cite{Stolzenwald:2018un}. Figure \ref{fig:blockcopygameparticipant} shows an example of a participant solving the puzzle. 
	
	For the data collection, 16 participants (7 females, $m_{age}$ = 25, \textit{SD} = 4) were recruited.
	Each completed one practice trial to get familiar with the procedure, followed by another three trials for data collection, where stock pieces and model pieces were randomised prior to execution. The pattern consists of 24 parts with an even count of the 4 types. 
	The task starts with 5 pre-completed pieces to increase the diversity of solving sequences leaving 19 pieces to be completed by the participant. That way, a total amount of 912 episodes of picking and dropping were recorded. 
	
	

	\subsection{User Intention Model}
	
	In the context of our handheld robot task, we define intention as the user's choice of which object to interact with next i.e. which stock piece to pick and on which pattern field to place it. 
	
	Based on our literature review, our modelling is guided by the following assumptions. 
	\begin{enumerate}[label=\textbf{A\arabic*}]
		\item An intended object attracts the users' visual attention prior to interaction. \label{A1}
		\item During task planning, the users' visual attention is shared between the intended object and other (e.g. subsequent) task-relevant objects.\label{A2}
	\end{enumerate}
	
	As a first step towards feature construction, the gaze information for an individual object was used to extract a visual attention profile (VAP) which is defined as the continuous probability of an object being gazed. Let $\vect{x_{gaze}}$ be the 2D point of intersection between the gaze ray and the TV screen surface and $\vect{x_{i}}$ the 2D position of the $i$-th object in the screen. Then the gaze position can be compared to each object using the Euclidean distance:
	\vspace{-0.5em}
	\begin{equation}
	d_i(t) = ||\vect{x_{gaze}}-\vect{x_{i}}||
	\end{equation}
	
	As a decrease of $d$ implies an increased visual intention, the distance profile can be converted to a visual attention profile (VAP) using the following equation:      
	\vspace{-0.3em}
	\begin{equation}    
	\vspace{-0.5em}
	P_{gazed,i}(t) = \exp(\frac{-d_i(t)^2}{2\sigma^2})
	\end{equation}

\begin{figure}[b]        
	\vspace{-1.011em}
	\centering
	\includegraphics[width=0.99\linewidth]{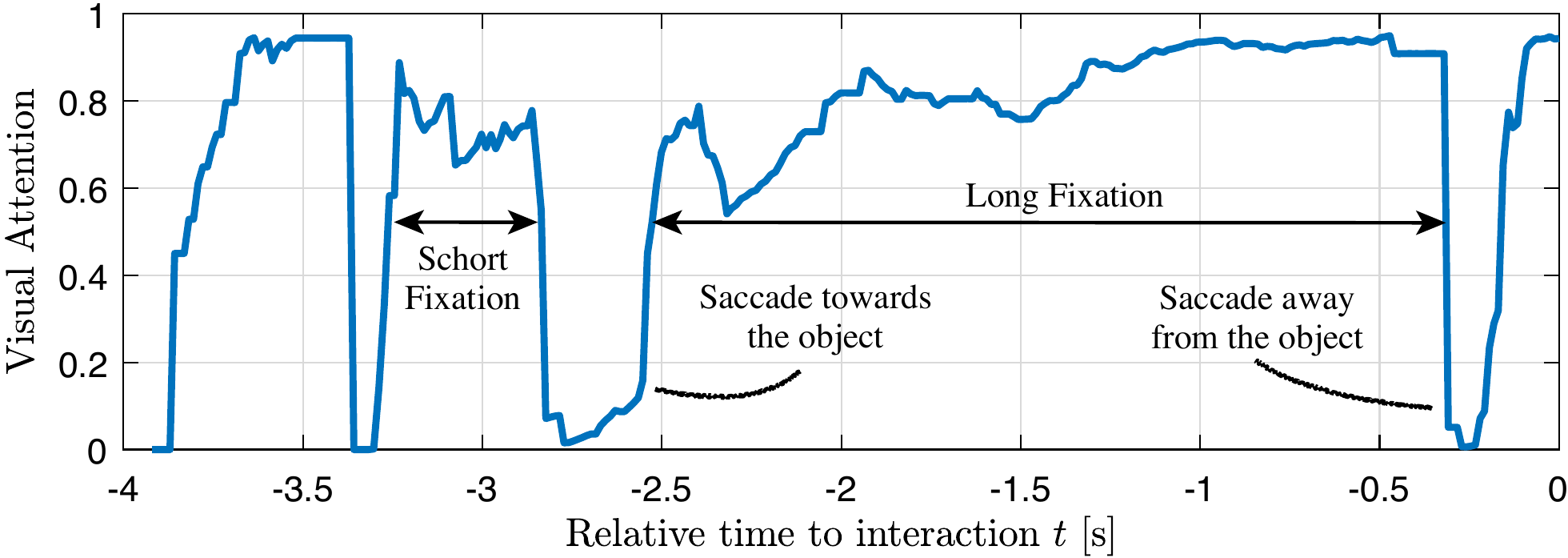}        
	\vspace{-1.5em}
	\caption{Illustration of changing visual attention over time within the anticipation window of the prediction model for an individual object.
	}
	\label{fig:visualattentionprofile}
	\vspace{0.7em}
\end{figure}

	Where $\sigma$ defines the gaze distance resulting in a significant drop of $P_{gazed}$ and it was set to \SI{60}{\mm} based on the pieces' size and tracking tolerance. The intention model uses the VAP of the last \SI{4}{\s} before the point in time of the prediction. Due to the data update frequency of \SI{75}{Hz} the profile is discretised into a vector of 300 entries (cf. example in figure \ref{fig:visualattentionprofile}).
	
	The prediction for picking and placing actions was modelled separately as they require different feature sets. As mentioned above, earlier studies about gaze behaviour during block copying \cite{Ballard:1995iy} and assembly \cite{Mennie:2006fo} suggest that the eye gathers information about both what to pick and where to place it prior to picking actions. For this reason, we combined pattern and stock information for picking predictions for each available candidate, resulting in the features selection:
	
	\begin{enumerate}[leftmargin=.42in]
		\item[$F_1$] The VAP of the object itself.
		\item[$F_2$] The VAP of the matching piece in the pattern. If there are several, the one with the maximum visual attention is picked. 
	\end{enumerate}

	This goes in line with our assumptions \ref{A1}, \ref{A2}. Both features are vectors of real numbers between 0 and 1 with a length of $n = 300$. For the prediction of the dropping location, \ref{A2} is not applicable as the episode finishes with the placing of the part hence why only $F_1$ (a vector with length $n = 300$) is used for prediction. Note that this feature contains information about fixation durations as well as saccade counts. 
	
	An SVM \cite{Hearst:1998ew} was chosen as a prediction model as this type of supervised machine learning model was used for similar classification problems in the past, e.g. \cite{Huang:2015iw}. We divided the sets of VAPs into two categories, one where the associated object was the intended object (labelled as \texttt{chosen = 1}) and another one for the objects that were not chosen for interaction (labelled as \texttt{chosen = 0}). Training and validation of the models were done through 5-fold cross validation \cite{Kohavi:1995wf}. 
	
	The accuracy of predicting the \texttt{chosen} label for individual objects is 89.6\% for picking actions and 98.3\% for placing. However, sometimes the combined decision is conflicting e.g when several stock pieces are predicted to be the intended ones. This is resolved by selecting the one with the highest probability $P($\texttt{chosen }$ = 1)$ in a one-vs-all setup \cite{Rifkin:2004vf}. This configuration was tested for scenarios with the biggest choice e.g. when all 4 stock parts (random chance = 25\%) would be a reasonable choice to pick or when the piece to be placed matches 4 to 6 different pattern pieces (random chance = 17-25\%). This results in a correct prediction rate of 87.9\% for picking and 93.25\% for placing actions when the VAPs of the time up to just before the action time is used.     

	\section{Results of Intention Modelling}
	Having trained and validated the intention prediction model for the case where VAPs range from $-4$ to 0 seconds prior to the interaction with the associated object, we are now interested in knowing to what extent the intention model predicts accurately at some time $t_{prior}$ prior to interaction. To answer this question, we extend our model analysis by calculating a $t_{prior}$-dependent prediction accuracy. Within a 5-fold cross validation setup, the \SI{4}{s}-anticipation window is iteratively moved away from the time of interaction and the associated VAPs are used to make a prediction about the subsequent user action using the trained SVM models. The validation is based on the aforementioned low-chance subsets, so that the chance of correct prediction through randomly selecting a piece would be $\leq25\%$. The shift of the anticipation window over the data set is done with a step width of 1 frame (\SI{13}{ms}). This is done for both the case of predicting which piece is picked up next as well as inferring intention concerning where it is going to be placed. For the time offsets $t_{prior}$ = 0, 0.5 and 1 seconds, the prediction of picking actions yields an accuracy $a_{pick}$ of 87.94\%, 72.36\% and 58.07\%. The performance of the placing intention model maintains a high accuracy over a time span of \SI{3}{s} with an accuracy $a_{place}$ of 93.25\%, 80.06\% and 63.99\% for the times $t_{prior}$ = 0, 1.5 and 3 seconds. In order to interpret these differences in performance, we investigated whether there is a difference between the mean duration of picking and placing actions. We applied a two-sample t-test and found that the picking time (mean = \SI{3.61}{s}, \textit{SD} = \SI{1.36}{s}) is significantly smaller than the placing time (mean = \SI{4.65}{s}, \textit{SD} = \SI{1.34}{s}), with $ p < 0.001, t = -16.12$. 
	
\begin{figure}[h]
		\centering
		\includegraphics[width=0.99\linewidth]{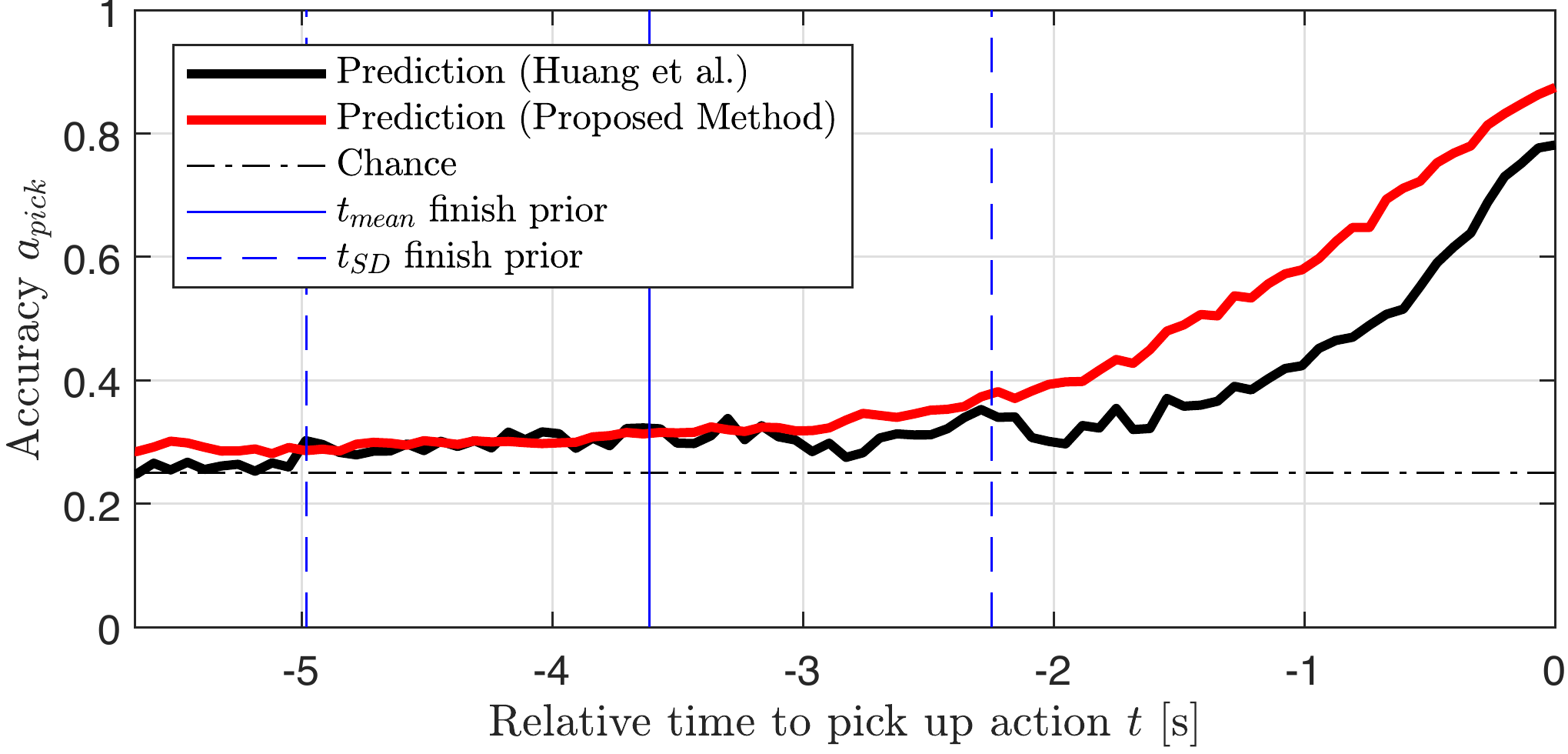}
		\caption{This diagram shows the performance of predicting pick up actions averaged over 912 samples for two models: our proposed model (red) and an SVM (black) which is based on the feature $F_{1}$ only,  as proposed by Huang et al. \cite{Huang:2015iw}. It can be seen how both models perform better than chance (dashed black) and predict the actions with increasing accuracy as the prediction time $t$ approaches the time of the action's execution $t = 0$. $t_{mean}$ (with temporal SD $t_{SD}$) is the mean time of completing the last block and hence the earliest meaningful time of predicting picking as a subsequent action.}
		\label{fig:pickuppredictionovertimecompare}
	\end{figure}

	As the prediction model of the picking actions implements the novel aspect of adding the VAPs of related objects, its comparison to existing methods is of particular interest. Figure \ref{fig:pickuppredictionovertimecompare} shows a comparison of our proposed model (where both features $F_{1}$ and $F_{2}$ are used) to the case where $F_{1}$ is the single basis for a prediction such as the model recently explored by Huang et al. \cite{Huang:2015iw}. It can be seen that both models well exceed the chance of picking randomly. Notably, the proposed model outperforms the existing one shortly after the subject ends the preceding move and presumably starts planning the next one. To further investigate the effect of the chosen model on the prediction performance, a two-factorial ANOVA was applied where the prediction time $t$ relative to the action and the model were set as the independent factors and the performance as dependent variable which reveals that the correct prediction rate of the proposed model is significantly higher ($p < 0.001$) than the one of the existing model.
	
	\subsection{Qualitative Analysis}
	For an in-depth understanding of how the intention models respond to different gaze patterns, we investigate the prediction profile i.e. the change of the prediction over time, for a set of typical scenarios.\\
	\subsubsection{One Dominant Type}
\begin{figure*}[t!]
		\vspace{0.5em}
		\centering
		\begin{subfigure}[t]{0.48\textwidth}
			\centering
			\includegraphics[width=0.99\linewidth]{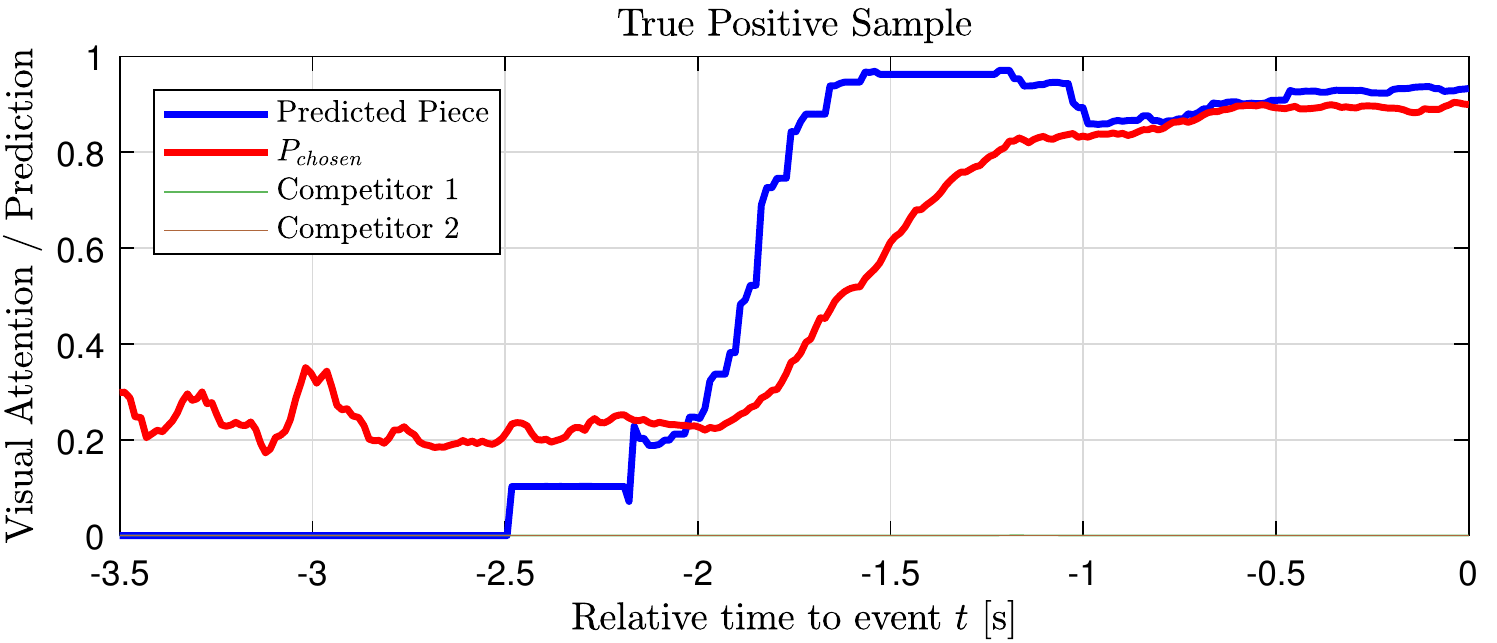}
			\caption{\scriptsize One piece receives most of the user's visual attention prior to placing}
			\label{fig:placingcorrecttruepositive1dominant3}
			\vspace{1em}
		\end{subfigure}
		%
		~ 
		\begin{subfigure}[t]{0.48\textwidth}
			\centering
			\includegraphics[width=0.99\linewidth]{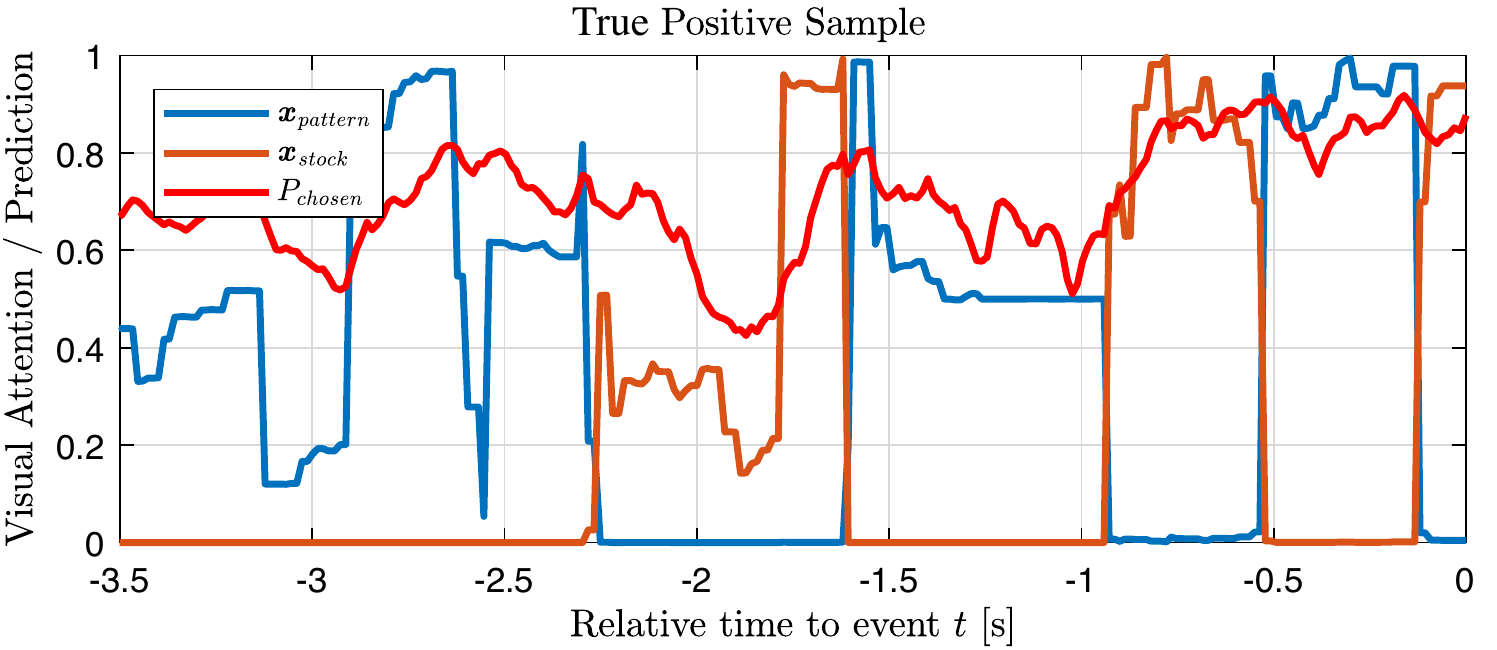}
			\caption{\scriptsize User gaze alters between stock piece and matching workspace location}
			\label{fig:pickupcorrecttuenegativethereandbackwideformat}
		\end{subfigure}%
		
		\caption{
			These diagrams show examples of correct predictions for \textit{one type dominant} samples. (a) shows, how long fixation times (blue) results into a high probability value (red) e.g. for a location to place a piece. Similarly, (b) shows, how the prediction model links the VIPs of related objects. The subject's gaze alters between two related objects e.g. a piece to pick up and a matching location to place it (cf. orange and blue VAPs) leading to a high probability estimation (red) for this piece being the user-intended one.            
		}
		\label{fig:one dominant}
	\end{figure*}

	A common observation was that the target object perceived most of the user's visual attention prior to interaction which goes in line with our assumption \ref{A1}. An example of these \textit{one type dominant} samples can be seen in figure \ref{fig:placingcorrecttruepositive1dominant3}. A subset of this category is the case where the user's eye gaze alters between the piece to pick and the matching place in the pattern i.e. where to put it (cf. figure \ref{fig:pickupcorrecttuenegativethereandbackwideformat}) which supports our assumption \ref{A2}.
	For the majority of these one type dominant samples both the picking and placing prediction models predict correctly.  \\
	
	\subsubsection{Trending Choice}
\begin{figure*}[h]
		\centering
		\begin{subfigure}[h]{0.48\textwidth}
			\centering
			\includegraphics[width=0.99\linewidth]{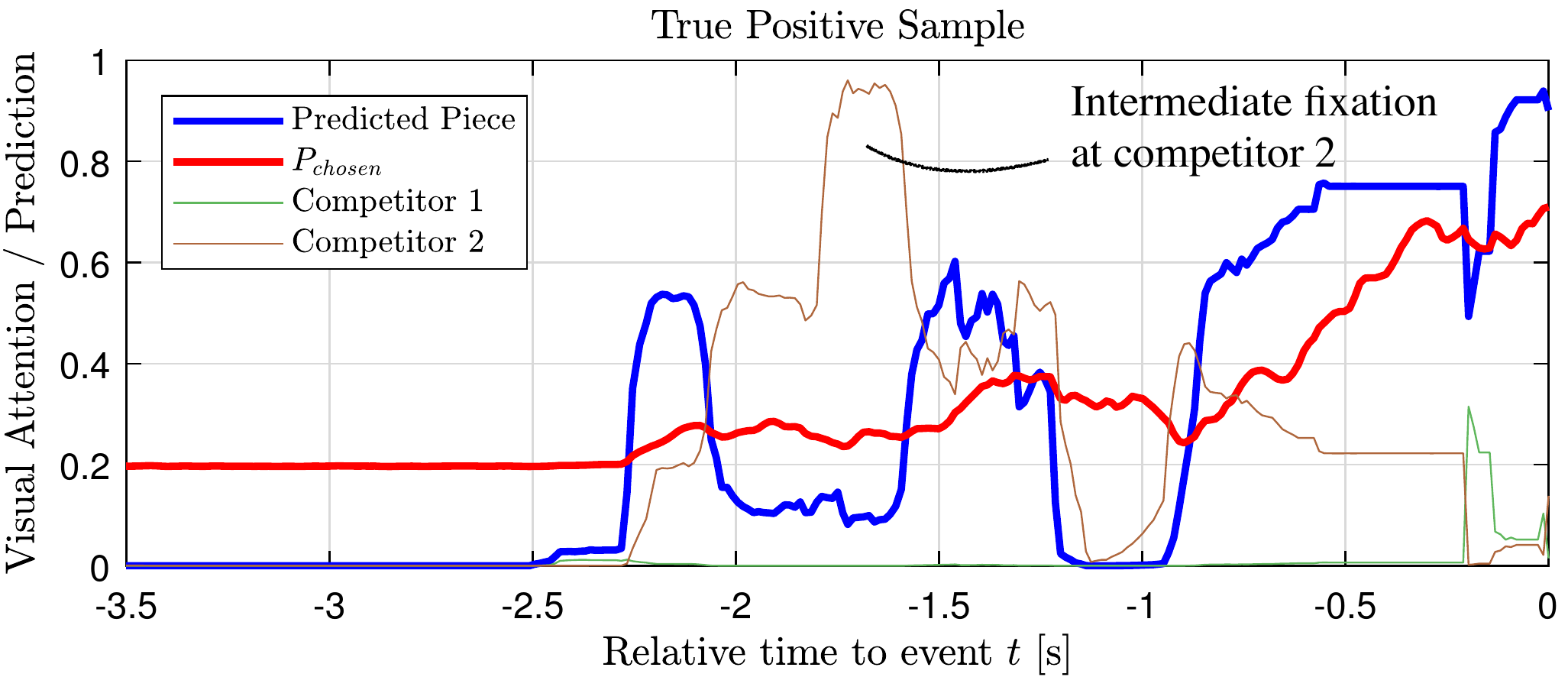}
			\label{fig:placingcorrecttruepositivetrendingchoice1}
		\end{subfigure}%
		~ 
		\begin{subfigure}[h]{0.48\textwidth}
			\centering
			\includegraphics[width=0.99\linewidth]{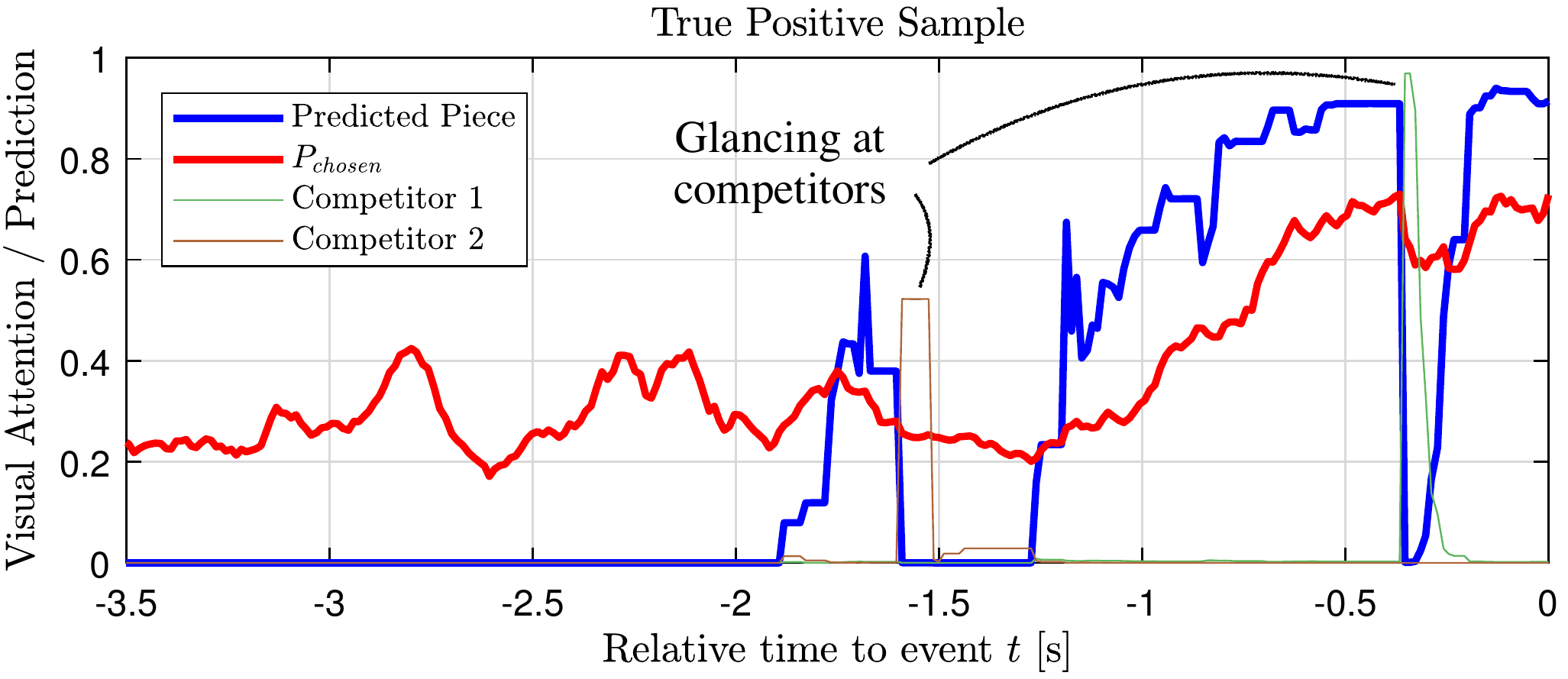}
			\label{fig:placingcorrecttruepositivetrendingchoice2}
		\end{subfigure}
		\vspace{-1em}
		\caption{These two examples illustrate how the visual attention (blue) of an object builds up during the user's decision process in which case the intention prediction (red) remains undecided ($P_{chosen} < 0.5$) for a longer time compared to the case where no competition receives fixations (cf. fig \ref{fig:one dominant}). }
		\label{fig:trending choice}
		\vspace{0.4em}
		\noindent\makebox[\linewidth]{\rule{\textwidth}{0.4pt}}
		\vspace{-2.5em}
	\end{figure*}

	While the anticipation time of the pick up prediction model lies within a second and is thus rather reactive, the placing intention model is characterised by a slow increase of likelihood during the task i.e. it shows a low-pass characteristic. Figure \ref{fig:trending choice} demonstrates that the model is robust against small attention gaps and intermediate glances at competitors, however, the model requires an increased time window to build up confidence.\\
	%
	
	\subsubsection{Incorrect Predictions}
	There is a number of reasons for an incorrect prediction. Most commonly, a close by neighbour received more visual attention and was falsely classified as the intended object. In other cases, it was impossible to predict the intended object using our model due to missing saccades towards it or faulty gaze tracking.

	\section{Discussion of Intention Modelling}
	In addressing research question \ref{Q1}, we proposed a user intention model based on gaze cues for the prediction of actions which was assessed in a pick and place task. As a novel aspect introduced through this study, the predictions are not only based on saccades and fixation durations of an individual object but also on those of related objects. In other words, assessing the attention on objects in the workspace helps to predict which piece outside the current workspace is needed next. When the subject turns his/her attention towards the piece, the model interprets this as a confirmation rather than the start of a selection process. This helps to cut the time required for the model to gather relevant gaze information and makes predictions more reliable than traditional models. 
	
	We showed that, within this task, the prediction of different actions has different anticipation times i.e. dropping targets are identified quicker than picking targets. This can partially be explained by the fact that picking episodes are shorter than placing episodes. But more importantly, we observed that users planned the entire pick-place cycle rather than planning picking and placing actions separately. This becomes evident through the qualitative analysis which shows altering fixations between the piece to pick and where to place it. That way, the placing prediction model is able to already gather information at the time of picking. 
	
	
	The proposed model allows predictions \SI{500}{ms} prior to picking actions (71.6\% accuracy) and \SI{1500}{ms} prior to dropping actions (80.06\% accuracy). These numbers are encouraging for testing the prediction model in a real-time application. Therefore, we proceed with an experimental study where the intention model is used for cooperative behaviour.

	\section{Intention Prediction Model Validation}
	\vspace{-1mm}
	In the second part of our study, we validate the proposed intention model for the case where it is used to control the robot's behaviour and motion. While the aforementioned experiments and analysis demonstrate that the intention model is capable of predicting users' short term goals while having full control over the robot's tip, it is unclear whether this is true for the case where the robot reacts to these predictions. For example, users might adapt their intention to the robot's plans just by seeing it moving towards a target which might differ from their initially intended move. That way, labelling the robot's predictions as being correct or incorrect in the same way as we did in the first study becomes invalid due to the lack of ground truth. For this reason, we propose to assess the intention model in an indirect way instead by observing users' reactions to the predictions with a focus on frustration. We hypothesise that a mismatch between the robot's and the user's plans would inflict user frustration and that frustration is reduced when the robot follows the true user intention compared to avoiding it. 
	
	\subsection{Intention Affected Robot Behaviour}
	
	For the experimental validation of the intention model, we used the aforementioned block copy task and introduced an assistive behaviour to the robot which is controlled based on the predictions of a user's intended subsequent move i.e. which piece the user wants to pick up next or at which location the user wants to drop it. We created 3 different behaviour modes: \textit{Follow intention}, \textit{Rebel} and \textit{Random}. For each, the robot retreats to a crouched position while there is a low probability for each available target. When the probability of the target with the highest probability reaches a threshold, the robot reacts as follows in the different modes:
	\begin{itemize}
		\item \textbf{Follow Intention: }\\ The robot moves towards the target with the highest predicted intention.
		\item \textbf{Rebel: } \\The robot avoids the target with the highest prediction and moves towards the target with the lowest predicted intention instead.
		\item \textbf{Random: }\\The robot moves towards a random target.
	\end{itemize}
	
	We set a maximum decision time of \SI{1.3}{s} after which the robot executes the above-mentioned behaviour for the rare case where no probability exceeds the threshold. This prevents the robot from getting stuck in the crouched position e.g. when there is a time gap in the gaze tracking stream. 
	
	\subsection{Experiment Execution}
	
	We recruited 20 new participants (6 females, $m_{age}$ = 26, \textit{SD} = 4) for the validation study of which 2 were later removed from the set for data analysis due to malfunctioning gaze tracking. Each was asked to first complete the task without the robot moving for familiarisation with the rules and the robot handling. This practice session was followed by 3 trials where, for each, the robot's behaviour was set to a different behaviour mode. The block pattern to complete as well as the order of the behaviour modes were randomised. Furthermore, 5 (out of 24) randomly chosen blocks were pre-completed to stimulate some diversity in solving strategies e.g. to prevent repeated line-by-line completion. 
	
	The participants were told to solve the trial tasks swiftly and that their performance was recorded. They did not receive any information about the behaviour modes but were told that the robot will move and try to help them with the task. Each trial was followed by the completion of a NASA Task Load Index (TLX) form \cite{Hart:1988ho} and \SI{3}{\min} resting time.

	\section{Results and Discussion: Model Validation}
	
\begin{figure}[h]
		\vspace{0.5em}
		\centering
		\includegraphics[width=0.85\linewidth]{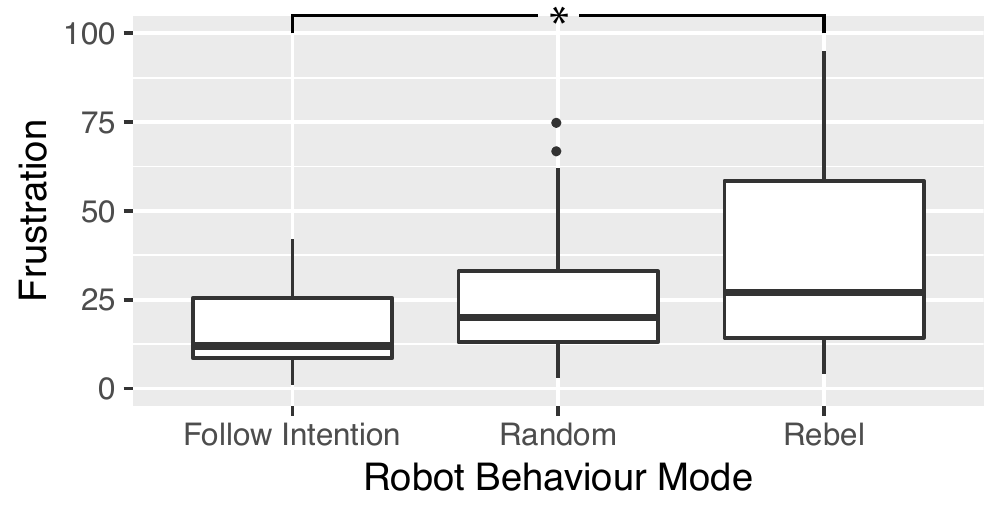}
		\vspace{-0.6em}
		\caption{Perceived frustration from the TLX results for each of the tested behaviour modes. The mean values of starred groups yield a significant difference (cf. table \ref{tab:frustration}).}
		\label{fig:intentionvalidationfrustration}
		\vspace{-1em}
	\end{figure}

	To determine the effect of the robot's behaviour mode on the subjects' frustration level, we performed an analysis of variance (ANOVA) with the mode as the independent variable and the frustration component of the TLX as a dependent variable. As the analysis yielded a significant effect ($p = .023$), it was further explored using post-hoc pairwise t-test with applied Bonferroni correction. The frustration mean for the \textit{Rebel} group was identified as being significantly higher than in the \textit{Follow Intention} group ($p = 0.19$). No significant mean differences were found when comparing the \textit{Random} group to the others. The results can be seen in table \ref{tab:frustration} and figure \ref{fig:intentionvalidationfrustration}.
	
	\begin{table}[b] \centering
		\vspace{-1.3em}
		\begin{tabular}{lll}
			& \textbf{Follow Intention}       & \textbf{Random}               \\ \cline{2-3} 
			\multicolumn{1}{l|}{\textbf{Rebel}}  & \multicolumn{1}{l|}{$p = .019$ *} & \multicolumn{1}{l|}{$p = .495$} \\ \cline{2-3} 
			\multicolumn{1}{l|}{\textbf{Random}} & \multicolumn{1}{l|}{$p = .469$}   & \multicolumn{1}{l|}{-}        \\ \cline{2-3} 
		\end{tabular}
		\caption{Bonferroni corrected \textit{p}-values of pairwise t-test results for the differences in mode depended frustration means. The starred value is significant ($p <.05$).}
		\label{tab:frustration}
		\vspace{1em}
	\end{table}
	
	We extended our analysis to both, the combined TLX results which serve as an indicator for perceived task load and the measured performance which is defined as the number of completed blocks per minute. However, an applied ANOVA did not yield an effect of the robot's behaviour mode, neither on the combined TLX nor on the performance. 
	
	\begin{figure*}[t]
		\vspace{0.5em}
		\centering
		\begin{subfigure}[t]{0.23\textwidth}
			\centering
			\includegraphics[width=0.99\linewidth]{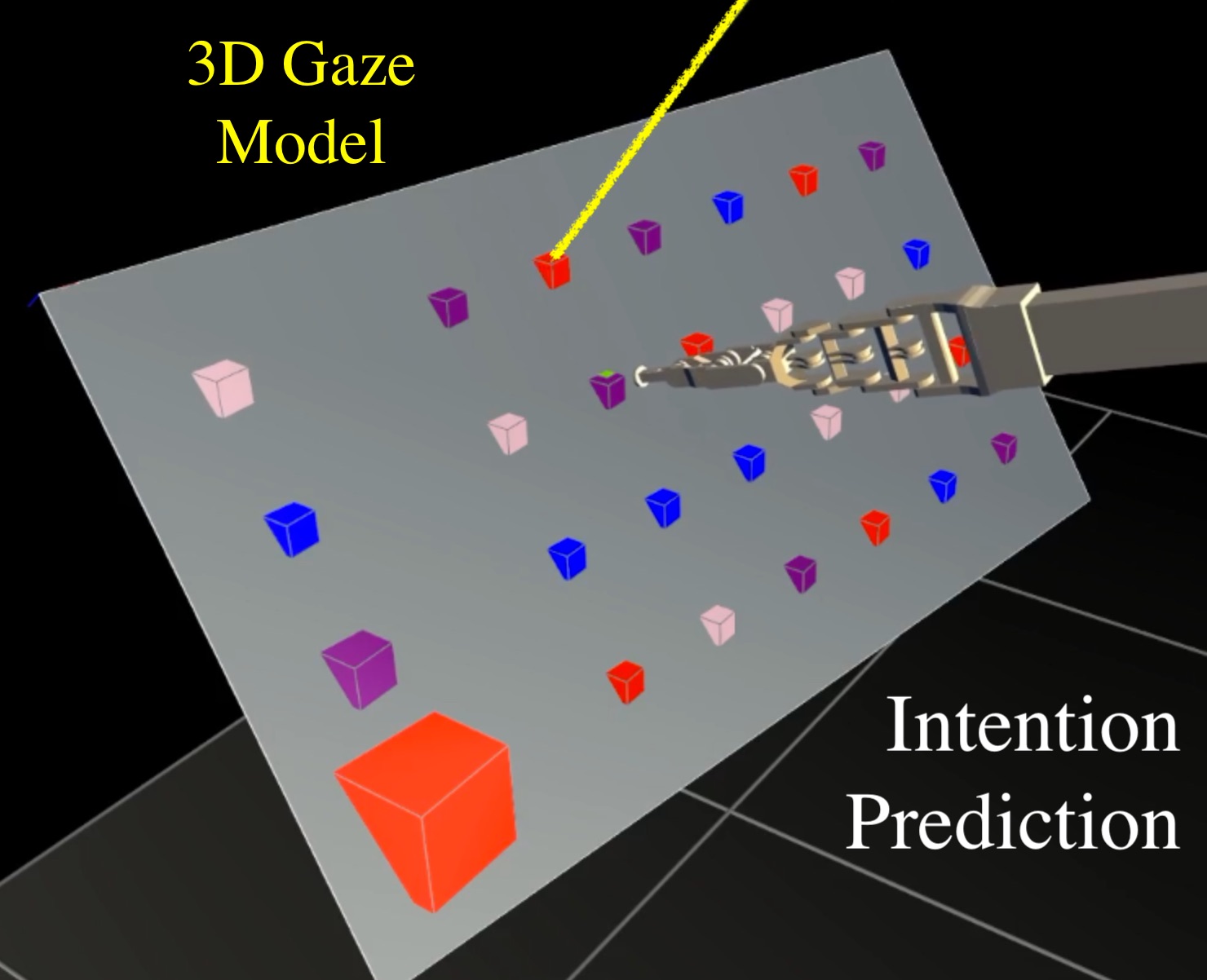}
			\caption{Prediction of the red piece during placing of the purple piece.}
			\label{fig:demofollowprediction-}
			\vspace{1em}
		\end{subfigure}
		~ 
		\begin{subfigure}[t]{0.23\textwidth}
			\centering
			\includegraphics[width=0.99\linewidth]{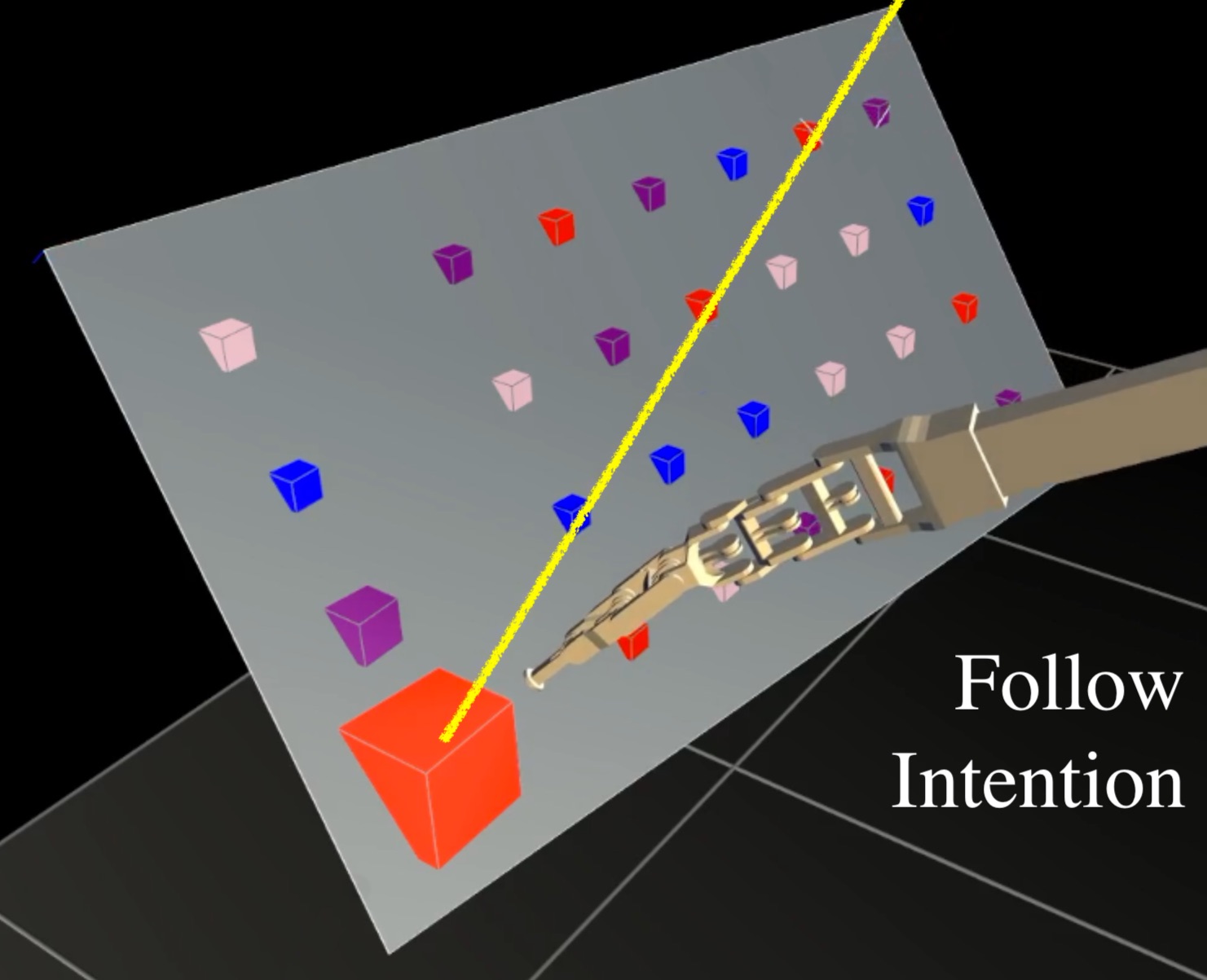}
			\caption{The robot's motion goes in line with the user's intention as it adapts its plans.}
			\label{fig:demofollowreaching}
		\end{subfigure}
		~
		\begin{subfigure}[t]{0.23\textwidth}
			\centering
			\includegraphics[width=0.99\linewidth]{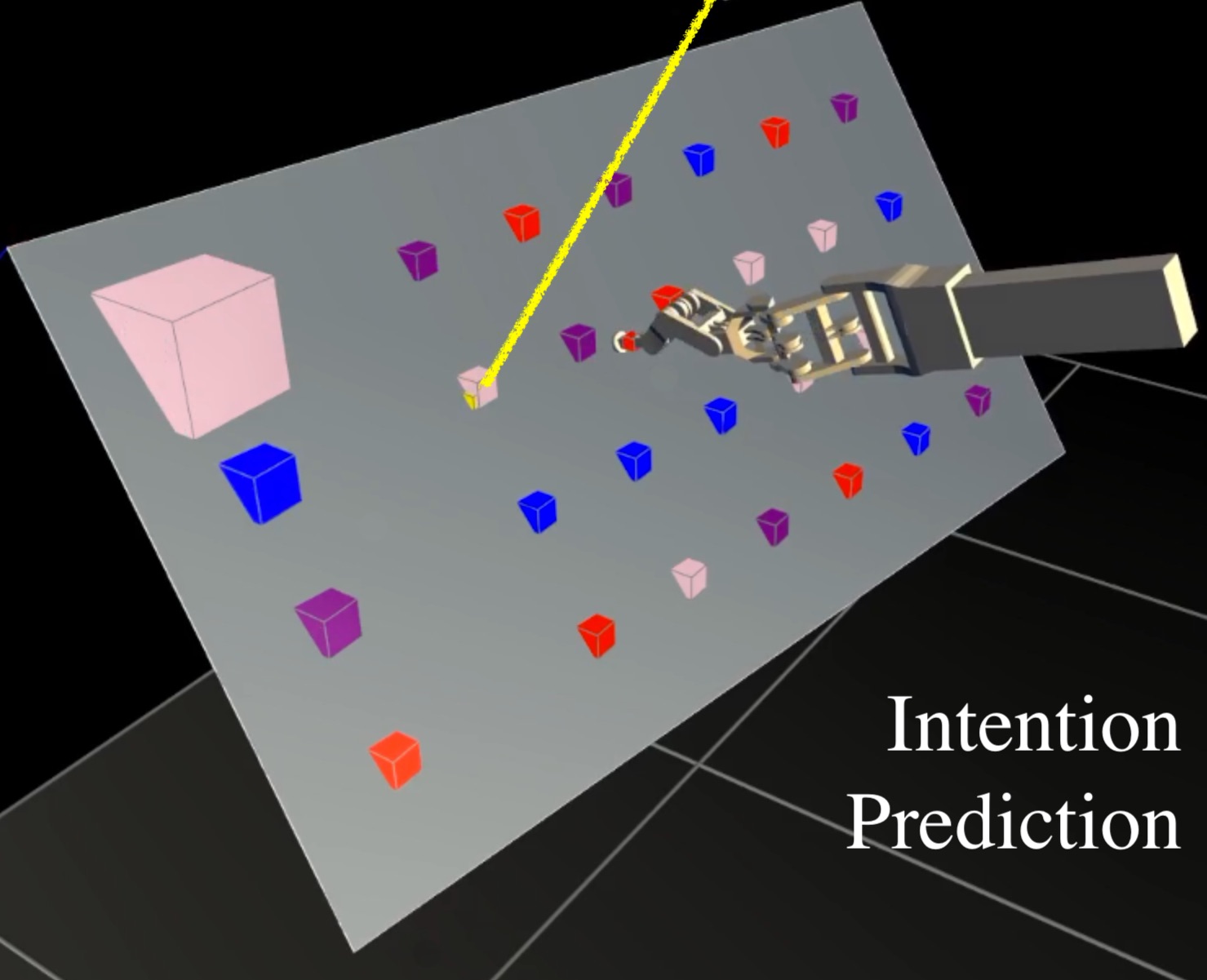}
			\caption{Prediction of the pink piece while placing the purple one.}
			\label{fig:demorebelprediction}
		\end{subfigure}
		~ 
		\begin{subfigure}[t]{0.23\textwidth}
			\centering
			\includegraphics[width=0.99\linewidth]{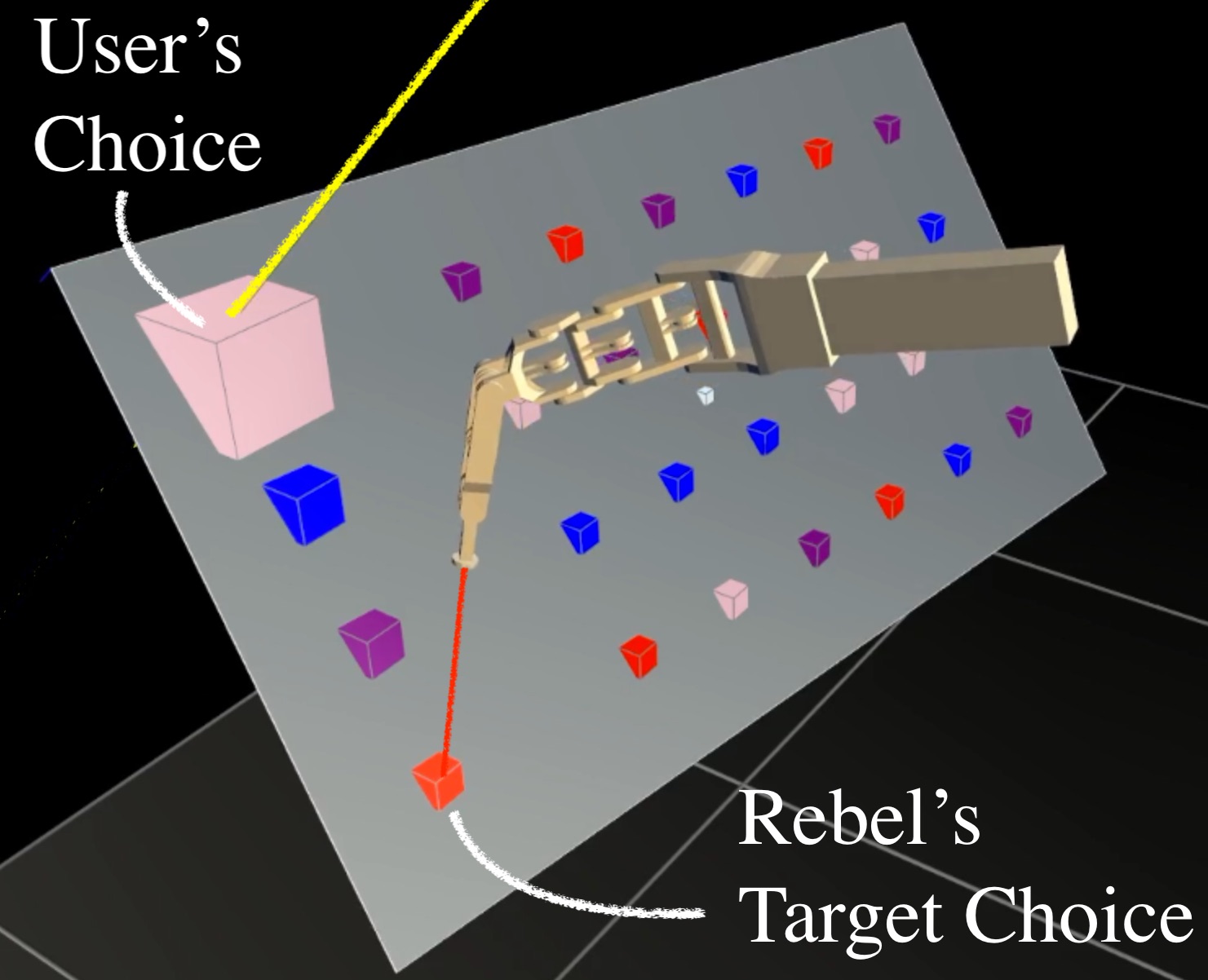}
			\caption{
				Avoiding user intent leads to a mismatch with the user's tactical motion.
			}
			\label{fig:demorebelreaching}
		\end{subfigure}
		\caption{These figures illustrate the systems' underlying intention estimation and how the different modes affect cooperation. The users' eye gaze model is represented as a yellow line while the estimated probability for a piece to be chosen by the user is indicated by its size. It can be seen how following the intention prediction assists the user with his/her choice (a,b) while avoiding the intended object (c,d) forces the user to adapt his/her plan to the robot's motion.}
		\label{fig:demo}
		\vspace{-0.5em}
	\end{figure*}
	
	As part of a qualitative review of the robot's behaviour we found that in the \textit{Rebel} mode, participants perform an increased number of corrective moves compared to the \textit{Follow Intention} scenario. Figure \ref{fig:demo} shows how the robot's aim matches the user's intention in the \textit{Follow Intention} mode whereas in the \textit{Rabel} example, the user rushes towards the intended aim but needs to correct his move as the robot aims for a different piece. 
	
	Some participants commented on the behaviour modes. The \textit{Follow Intention } mode was often preferred (e.g. \enquote{I liked being in charge and the robot was  helpful} and \enquote{The robot followed my decisions}) whereas the \textit{Random} mode lead to irritation in some users (e.g. \enquote{First I thought it would go where I wanted but then it started moving in an unpredictable way}). For the \textit{Rebel} mode, we observed divergent reactions. While some subjects struggled because of the mismatch between the robot's motion and their plans, others started following the robot's lead. This was also reflected in the comments e.g. \enquote{Now the robot does its own thing, I don't like it} versus \enquote{It was easier because I did not have to think much}.    
	\\
	
	\vspace{-0.5em}
	The observed difference in frustration ratings between the mode where the robot supports the user's predicted intention versus avoiding it is evidence for most of the intention predictions matching the true intention. With regards to \ref{Q2}, our interpretation of the results is that during the \textit{Follow Intention} trials, the robot did follow the users' preferred sequence rather than the users adapting it to the robotic motion which validates the proposed intention model and its application in assisted reaching. 
	
	The fact that the mean frustration for the \textit{Random} mode lies between the other two modes is expected given their effect on frustration outlined above. However, the effect is too subtle to be compared to random motion and the sample size too small for a reliable distinction.
	
	Our analysis furthermore shows that user frustration is more sensitive to the robot's intention prediction than perceived task load or performance. We suggest that robotic systems should follow user intention when there are subtasks with similar priorities for enhanced cooperation.
	\vspace{-0.8em}    

	\section{Conclusion}
	We investigated the use of gaze information to infer user intention within the context of a handheld robot. A pick and place task was used to collect gaze data as a basis for an SVM-based prediction model. Results show that, depending on the anticipation time, picking actions can be predicted with up to 87.94\% accuracy and dropping actions with an accuracy of 93.25\%. Furthermore, the model allows action anticipation \SI{500}{ms} prior to picking and \SI{1500}{ms} prior to dropping. We show that merging gaze information with respect to objects that are linked to the same task in a single model helps to increase the prediction performance.
	
	The developed intention model can be used to make predictions in real-time enabling the robot to align its plans to the user's preferred goals making it a cooperative tool for complex tasks. 
	
	The proposed model performs particularly well for tasks where several objects connect to the same subtasks. This opens its applicability to other tasks in assembly and assisted living.\\
	
	\vspace{-0.8em}
	{\bf Acknowledgements} To the German Academic Scholarship Foundation and UK's EPSRC. Opinions are the ones of the authors and not of the funding organisations.
	\vspace{-0.9em}
	
	%
	

	\bibliographystyle{unsrt} 
	\bibliography{references}

	
	

\end{document}